\documentclass{article}
\usepackage{spconf,amsmath,epsfig}
\usepackage{hyperref}       
\usepackage{url}            
\usepackage{booktabs}       
\usepackage{amsfonts}       
\usepackage{nicefrac}       
\usepackage{microtype}      
\usepackage{xcolor}         
\usepackage{graphicx}
\usepackage{algorithmic,algorithm}
\usepackage{multirow}
\usepackage{diagbox}
\usepackage{subfigure}

\let\OLDthebibliography\thebibliography
\renewcommand\thebibliography[1]{
  \OLDthebibliography{#1}
  \setlength{\parskip}{0pt}
  \setlength{\itemsep}{0pt plus 0.3ex}
}

\begin{document}\sloppy

\def\x{{\mathbf x}}
\def\L{{\cal L}}

\title{Improving Neural ODEs via Knowledge Distillation}
%
\name{Haoyu Chu, Shikui Wei, Qiming Lu, Yao Zhao}
\address{Institute of Information Science, Beijing Jiaotong University, Beijing 100044, China}
\maketitle
\begin{abstract}
Neural Ordinary Differential Equations (Neural ODEs) construct the continuous dynamics of hidden units using ordinary differential equations specified by a neural network, demonstrating promising results on many tasks. However, Neural ODEs still do not perform well on image recognition tasks. The possible reason is that the one-hot encoding vector commonly used in Neural ODEs can not provide enough supervised information. We propose a new training based on knowledge distillation to construct more powerful and robust Neural ODEs fitting image recognition tasks. Specially, we model the training of Neural ODEs into a teacher-student learning process, in which we propose ResNets as the teacher model to provide richer supervised information. The experimental results show that the new training manner can improve the classification accuracy of Neural ODEs by 24\% on CIFAR10 and 5\% on SVHN. In addition, we also quantitatively discuss the effect of both knowledge distillation and time horizon in Neural ODEs on robustness against adversarial examples. The experimental analysis concludes that introducing the knowledge distillation and increasing the time horizon can improve the robustness of Neural ODEs against adversarial examples.
\end{abstract}
\section{Introduction}
\label{sec:intro}

Neural ODEs proposed by~\cite{chen2018neural} aimed at constructing continuous deep neural networks by using ordinary differential equations specified by a neural network. Neural ODEs allow training continuous deep neural networks in an end-to-end manner, which have been successfully applied in numerous tasks, such as density estimation~\cite{grathwohl2018ffjord, onken2020ot}, time-series modeling~\cite{jia2019neural, kidger2020neural}, physics-based models~\cite{greydanus2019hamiltonian, chen2020learning} and some others~\cite{gupta2020neural, chen2020mri, giannone2020real}. Besides, the robustness of Neural ODEs to adversarial attacks has also been discussed in several literatures~\cite{huang2020adversarial, yan2019robustness, carrara2019robustness}. Although Neural ODEs show great potential in handling various learning tasks, they still do not show their effectiveness on image recognition tasks. Dupont et al.~\cite{dupont2019augmented} have reported that the accuracy of Neural ODEs on the CIFAR10 test set is only 53.7\%$\pm$0.2, which is relatively lower than the discrete version of deep learning models like ResNet~\cite{he2016deep}. The only successful attempt is to apply Neural ODEs to handwriting recognition~\cite{chen2018neural}. The reason for conflict lies in the complexity of image datasets. As is well known, the MINST dataset contains 70000 28$\times$28 black and white images representing the digits zero through nine, in which the scene of the image is quite simple.
In contrast, the CIFAR10 dataset consists of 60000 32$\times$32 color images in 10 classes, in which the scene of the image is more complex. It is not easy for Neural ODEs to handle some complex image recognition tasks. According to our analysis, the possible reason lies in that Neural ODEs lack enough supervised information to train networks for complex tasks. In existing Neural ODEs, the supervised information is commonly provided by a so-called one-hot encoding vector. Since the one-hot encoding vector is a hard label and cannot offer richer supervised information, Neural ODEs fail to find a suitable mapping from features to labels. Therefore, it is reasonable to improve the Neural ODEs by providing richer supervised information. The critical issue is how to obtain richer supervised information.

Motivated by knowledge distillation, we attempt to design a new training manner that captures richer supervised information. Knowledge distillation, originally proposed by \cite{bucilua2006model} and further developed by Hinton et al.~\cite{hinton2015distilling}, compress the knowledge of a large and computational expensive model to a single computational efficient neural network. Moreover, Papernot et al.~\cite{papernot2016distillation} showed that adversarial examples could be resisted based on knowledge distillation. However, they chose two identical models as the teacher model and the student model. It is valuable to discuss the effectiveness of knowledge distillation against adversarial examples in Neural ODEs.  

In this paper, we propose a new training manner for Neural ODEs, in which the teacher-student mechanism is employed to extract the knowledge learned by the discrete version of deep neural networks. The obtained richer supervised information can assist the training of Neural ODEs for more complex image recognition tasks. In particular, we treat ResNets as the teacher models. We experimentally demonstrate that sufficient supervised information can significantly improve Neural ODEs on complex image recognition tasks. The accuracy of Neural ODEs can reach up to 88.03\% on CIFAR10 and 93.53\% on SVHN. In addition to performance-boosting, we also quantitatively discuss the robustness of Neural ODEs trained by knowledge distillation against adversarial examples. The experimental results show that introducing the knowledge distillation indeed improves the robustness of Neural ODEs. Furthermore, the effect of time horizon on robustness is also analyzed qualitatively and quantitatively. A surprising conclusion is that increasing the time horizon of Neural ODEs can remarkably improve the robustness of Neural ODEs, and the robustness benefits from the time horizon rather than obfuscated gradients claimed by~\cite{huang2020adversarial}. 

\section{Method}

In this section, we first introduce Neural ODEs, analyze the possible reason why Neural ODEs have natural robustness against adversarial examples, and then introduce knowledge distillation. Finally, we propose knowledge distillation for training Neural ODEs and present the corresponding pseudocode to illustrate this procedure.

\subsection{Background}
\subsubsection{Neural ODEs}
Given an input $\mathbf{y}_t$, the residual block proposed by~\cite{he2016deep} can be described as $\mathbf{y}_{t+1}=\mathbf{y}_t+hf(\mathbf{y}_t,\boldsymbol{w})$ with step size $h=1$. The residual block has been considered as a particular case of Euler's method \cite{weinan2017proposal, lu2018beyond}. When taking smaller steps and add more layers, which in the limit, Neural ODEs parameterize the continuous dynamics of hidden units using ordinary differential equations specified by a neural network
\begin {equation} \label{equ:neural_odes}
\lim_{h\rightarrow0}\frac{\mathbf{y}_{t+h}-\mathbf{y}_t}{h}=\frac{d\mathbf{y}}{dh}=f(t,\mathbf{y},\boldsymbol{w}),
\end {equation}
where $\boldsymbol{w}$ is the trainable parameter in Neural ODEs.

Therefore, the analytical solution of Eqn.\ref{equ:neural_odes} can be given by 
\begin {equation} \label{equ:analytical_solution}
\mathbf{y}_{t_{1}} = \mathbf{y}_{t_{0}} + \int_{t_{0}}^{t_{1}}f(t,\mathbf{y},\boldsymbol{w})\mathrm{d}t,
\end {equation}
where $[t_{0}, t_{1}]$ represents the time horizon for solving ordinary differential equations. 

In common configurations, Neural ODEs apply the Dopri5 method~\cite{dormand1980family} as the ODE solver, which can adaptively adjust the step size until that the desired tolerance is reached. At each step, two different approximations (the 4th order and 5th order Runge–Kutta methods) for the solution are made and compared. If the two answers are in close agreement, the approximation is accepted. If the two answers do not agree to a specified accuracy, the step size is reduced. If the answers agree to more significant digits than required, the step size is increased. This process is detailed in the supplementary material. Because of this property, Huang et al.~\cite{huang2020adversarial} claimed that Dopri5 might lead to obfuscated gradients via large error tolerance with adaptive step size choice, which fails the gradient-based attacks like PGD. However, in this paper, we attempt to experimentally show that the robustness of Neural ODEs benefits from the time horizon rather than obfuscated gradients, and increasing the time horizon can remarkably improve the robustness of Neural ODEs. 

\begin{algorithm}[t]
\caption{Knowledge distillation for Neural ODEs}
\label{alg:algorithm}
\begin{algorithmic}[1]
\REQUIRE A training set $\boldsymbol{x}$ with hard targets $\boldsymbol{y}$;
a temperature $T$;
a factor $\lambda$;
a trained ResNet model $M_{R}$;
the number of epochs $E$.
\ENSURE The Neural ODEs model $M_{N}$.
\STATE $\boldsymbol{l}^{R}=M_{R}(\boldsymbol{x})$; \\
\hfill \%Compute the logits on the ResNet.
\STATE $\boldsymbol{s}^{R}=softmax(\boldsymbol{l}^{R}/T)$; \\
\hfill \%Compute the soft targets on the ResNet.
\FOR{$i=0$ to $E-1$} 
    \STATE Initial the weight $\boldsymbol{w}_{0}$ in the model $M_{N}$ by He initialization~\cite{he2015delving}. 
    \STATE $\boldsymbol{l}^{N}=M_{N}(\boldsymbol{x})$; \\
    \hfill \%Compute the logits on Neural ODEs.
    \STATE $\boldsymbol{h}^{N}=softmax(\boldsymbol{l^{N}})$; \\
    \hfill \%Compute the hard targets on Neural ODEs.
    \STATE $\boldsymbol{s}^{N}=softmax(\boldsymbol{l^{N}}/T)$; \\
    \hfill \%Compute the soft targets on Neural ODEs.
    \STATE $\mathcal{L}_{SL}=\mathcal{L}_{cross-entropy}(\boldsymbol{h}^{N}, \boldsymbol{y})$ \\ \hfill \%Compute the standard learning loss.
    \vspace{0.5ex}
    \STATE $\mathcal{L}_{KD}=D_{KL}(log(\boldsymbol{s}^{N})\Vert \boldsymbol{s}^{R})$ \\
    \hfill \%Compute the knowledge distillation loss.
    \vspace{0.5ex}
    \STATE $\mathcal{L}=(1-\lambda)\mathcal{L}_{SL}+\lambda T^{2} \mathcal{L}_{KD}$ \\
    \hfill \%Compute the loss for training Neural ODEs.
    \STATE $\boldsymbol{w}_{i}=\boldsymbol{w}_{i}-\nabla_{\boldsymbol{w}_{i}}\mathcal{L}$ \\
    \hfill \%Update the weight of Neural ODEs.
\ENDFOR
\RETURN $M_{N}$.
\end{algorithmic}
\end{algorithm}

\subsubsection{Knowledge Distillation}
The key concept in knowledge distillation proposed by~\cite{hinton2015distilling} is soft target. Soft target is got by using a softmax output layer that converts the logits, $\mathbf{l}_{i}$, computed for each class into a probability, $\mathbf{s}_{i}$, by comparing $\mathbf{l}_{i}$ with the other logits. The computation of the soft target can be formalized by 
\begin{equation} \label{equ:kd}
\mathbf{s}_{i}=\frac{exp({\mathbf{l}_{i}/T)}}{\sum_{j}{exp(\mathbf{l}_{j}/T)}},
\end{equation}
where $T$ is the temperature. 

The procedure of knowledge distillation is first to obtain the soft targets on the teacher model. The loss function of the student model is a weighted sum of the cross-entropy loss of the soft targets and the cross-entropy loss of the hard targets. The student model is trained in this way so that the student model can learn from the knowledge of the teacher model.

\subsection{Distilling the Knowledge from ResNet to Neural ODEs}

Inspired by knowledge distillation, we propose a new training manner for Neural ODEs. The training of Neural ODEs is modeled into a teacher-student learning process, in which ResNet is treated as the teacher to provide richer supervised information. We first train the teacher ResNet model, then use both the hard targets and the soft targets got by ResNet to train Neural ODEs. The detailed procedure is summarized in Algorithm~\ref{alg:algorithm}. In our experimental settings, we use 5 teacher ResNet models (ResNet20, ResNet32, ResNet44, ResNet56, ResNet110) to individually get the soft targets. We use ResNet as the teacher model because ResNet is considered as a discrete version of the Neural ODEs, and it is the most similar model to the Neural ODEs. 

\section{Experimental Setup} \label{section::experimental_setup}

\subsection{The Network Architecture of Neural ODEs}

We parameterize the continuous dynamics of hidden units by a 4 layers convolutional neural network with 3$\times$3 kernel size, we set the time horizon as [0, 1] or [0, 100]. We choose such network architecture of Neural ODEs for two reasons: (1) Knowledge distillation is considered as a network compression method. A small student model can mimic the comprehensive teacher model by using the cross-entropy loss of the soft targets. (2) The depth of the Neural ODEs is not as same as the convolutional neural networks. For now, it is not clear how to define the depth of the Neural ODEs. Chen et al.~\cite{chen2018neural} remarked that a related quantity is the number of evaluations of the hidden state dynamics required. In this view, the depth of Neural ODEs is usually larger than the depth of ResNet architectures, because the number of evaluations is quite large in practical computations.
\subsection{Impementation}

We use PyTorch~\cite{paszke2017automatic} framework for the implementation. As a pre-processing step, we use normalization and randomized horizontal crop for both datasets. 

For optimization, we use Stochastic Gradient Descent (SGD) with the moment of 0.9 for ResNets and Neural ODEs with the time horizon [0, 1]. In our experiments, we find that the end time for the training of Neural ODEs using knowledge distillation can not be over 5 ($t_{1}$ in Eqn.\ref{equ:analytical_solution}). A larger end time, such as 2, yields poor knowledge distillation results than the end time 1 when using SGD as the optimization algorithm. Thus, Neural ODEs with the time horizon [0, 100] are trained by the Adam optimization algorithm~\cite{kingma2014adam}. Besides, all of the Neural ODEs without using knowledge distillation are trained by the Adam optimization algorithm. 

For training the teacher ResNet models, the training epochs are set to 300 for CIFAR10 and 40 for SVHN. The mini-batch size is set to 128 for both datasets. The initial learning rate is set to 0.1 (0.01 for ResNet110). We decrease the learning rate by 1/10 of the initial learning rate on 1/2 epochs and decrease the learning rate by 1/100 of the initial learning rate on 3/4 epochs. For training the student Neural ODEs models, the training epochs are set to 200 for CIFAR10 and 40 for SVHN. The initial learning rate is set to 0.001 with the same learning rate decreasing method above. We set $\lambda$ in Algorithm~\ref{alg:algorithm} to 0.9. 

\section{Experimental Results}
In this section, we evaluate knowledge distillation for Neural ODEs and investigate why Neural ODEs are robust to adversarial examples. Throughout this section, we use Neural ODEs($t_{1}$) or NODEs($t_{1}$) to represent the Neural ODEs with time horizon [0, $t_{1}$] and DATASETS(adversarial method) to represent adversarial input images with a certain adversarial attack method.

\begin{table*}[t]
\caption{The classification accuracy of Neural ODEs without using knowledge distillation and ResNets on CIFAR10 and SVHN.}
\centering
\begin{tabular}{cccccccc}
\toprule
\diagbox{Datasets}{Model}&NODEs(1)&NODEs(100)&ResNet20&ResNet32& ResNet44 & ResNet56 & ResNet110 \\
\midrule
CIFAR10 & 63.45 & 62.38 &89.09  & 88.75  & 90.32 &  89.95 & 88.26\\
\midrule
SVHN& 88.36 & 88.68 & 91.68 & 90.96  & 93.20  & 94.58  & 92.54  \\
\bottomrule
\end{tabular}
\label{tab:node_resnet_original}
\end{table*}

\begin{table*}[t]
\caption{The classification accuracy of the student Neural ODEs(1) model on CIFAR10 and SVHN. We fix the number of channels to 512 and fine-tune the temperature $T$.}
\centering
\begin{tabular}{ccccccc}
\toprule
Datasets &\diagbox{$T$}{Teacher}&ResNet20&ResNet32& ResNet44 & ResNet56 & ResNet110 \\
\midrule
\multirow{3}{*}{CIFAR10} & 3                                  &    86.72      &    86.42      &    86.08      &   86.32       &   86.47     \\
                         & 10                                 & 87.36    &  87.75   & \textbf{88.03}    & 86.94    & 87.80     \\
                         & 20                                 & 87.96    & 87.79   &  87.10        & 87.01    &  87.41         \\
\midrule
\multirow{3}{*}{SVHN}    & 3                                  &  93.46   &  92.89   &  93.12 &  93.02   & \textbf{93.53}    \\
                         & 10                                 &  93.12    &   92.74  &   92.46   &   92.74   &  93.14    \\
                         & 20                                 &  92.77   & 92.24  &  92.22 &  92.67  & 93.18    \\
\bottomrule
\end{tabular}
\label{exp:kd_finetune_t}
\end{table*}

\begin{figure*}[t]
\begin{center}
\subfigure[The evaluation on CIFAR10 under PGD.]{\includegraphics[width=7.9cm]{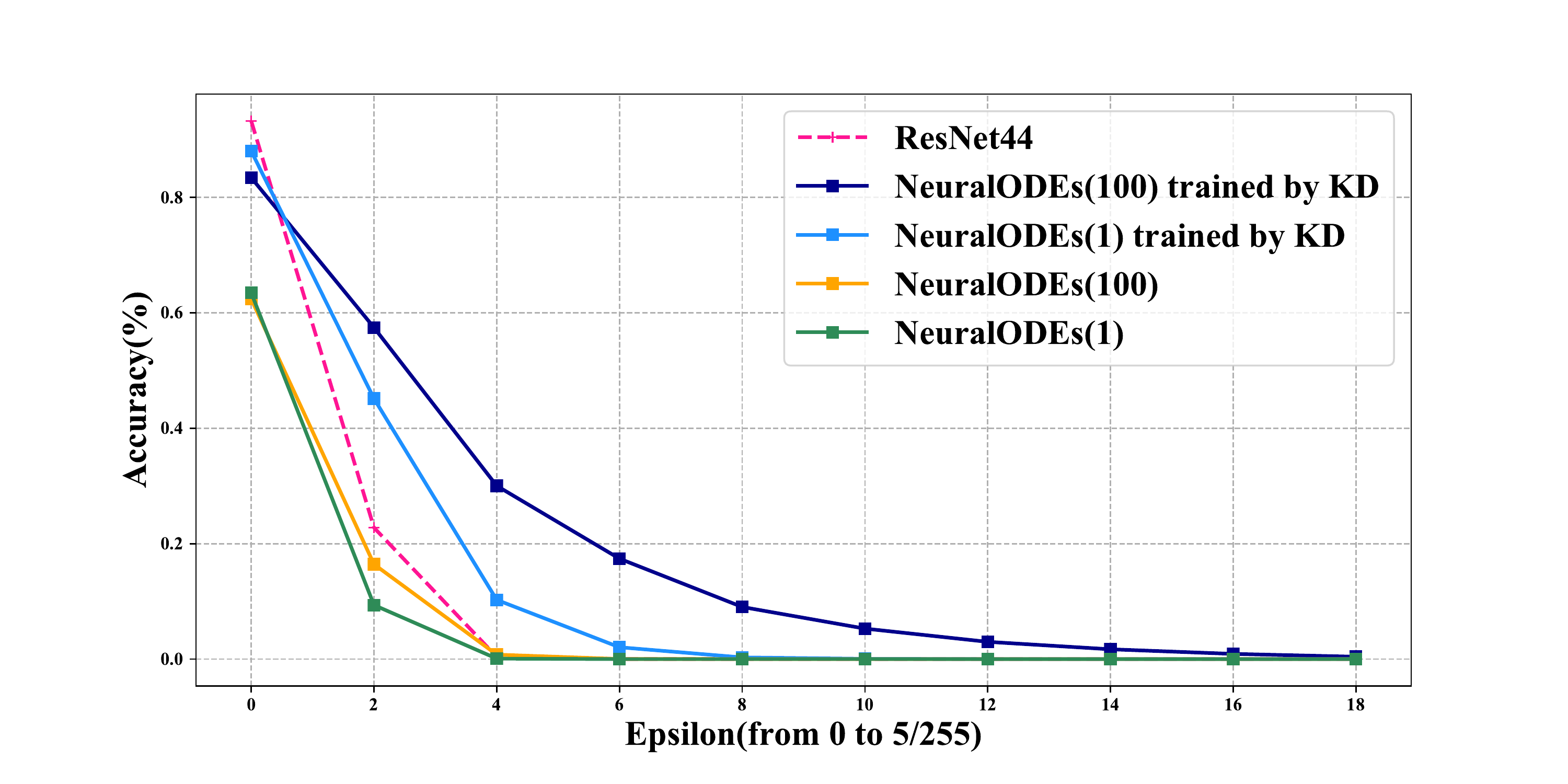}
  \label{fig:subfigure1}}
\quad
\subfigure[The evaluation on SVHN under PGD]{\includegraphics[width=7.9cm]{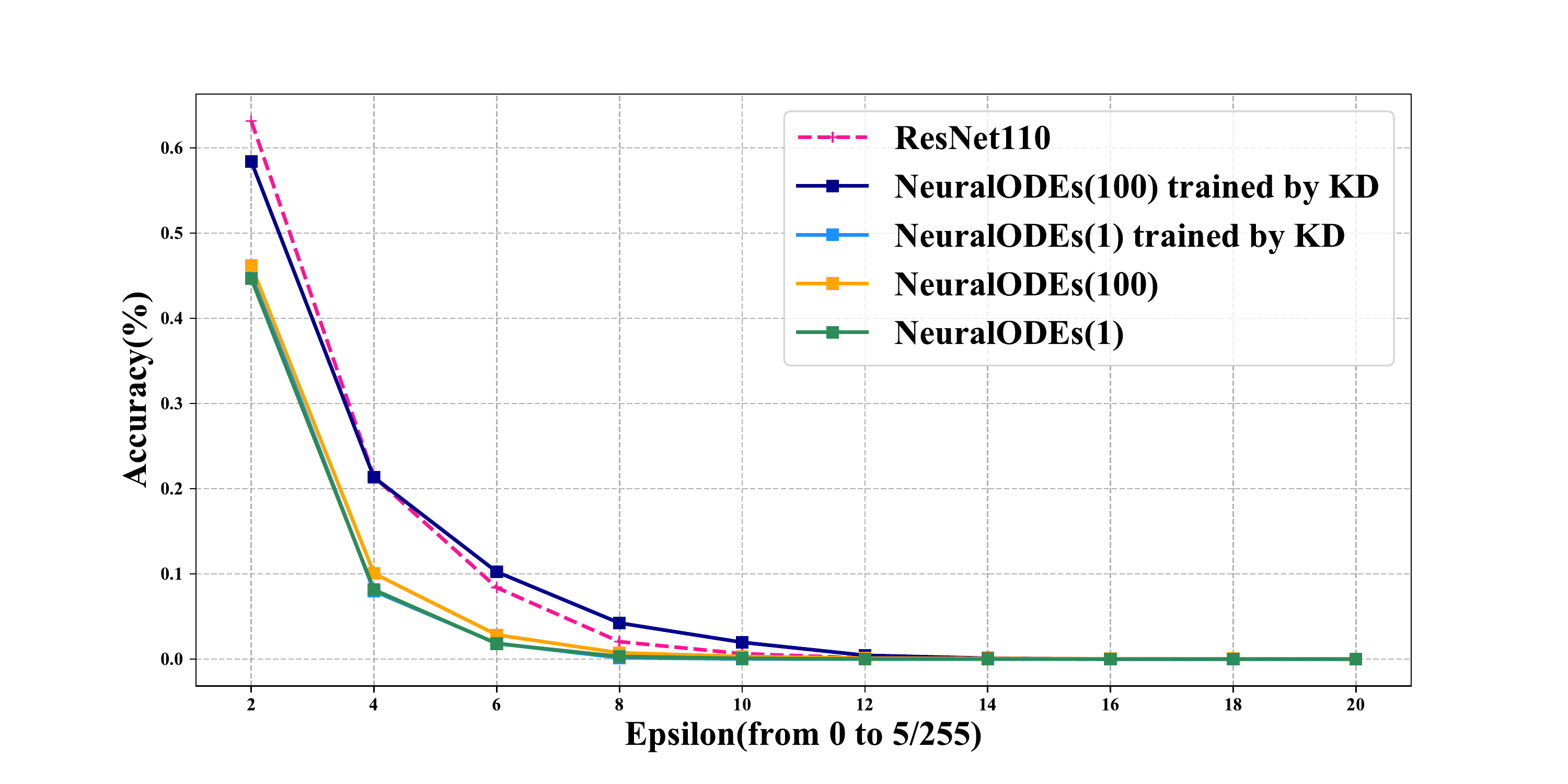}
 \label{fig:subfigure2}}
 
\subfigure[The evaluation on CIFAR10 under MI-FGSM.]{\includegraphics[width=7.9cm]{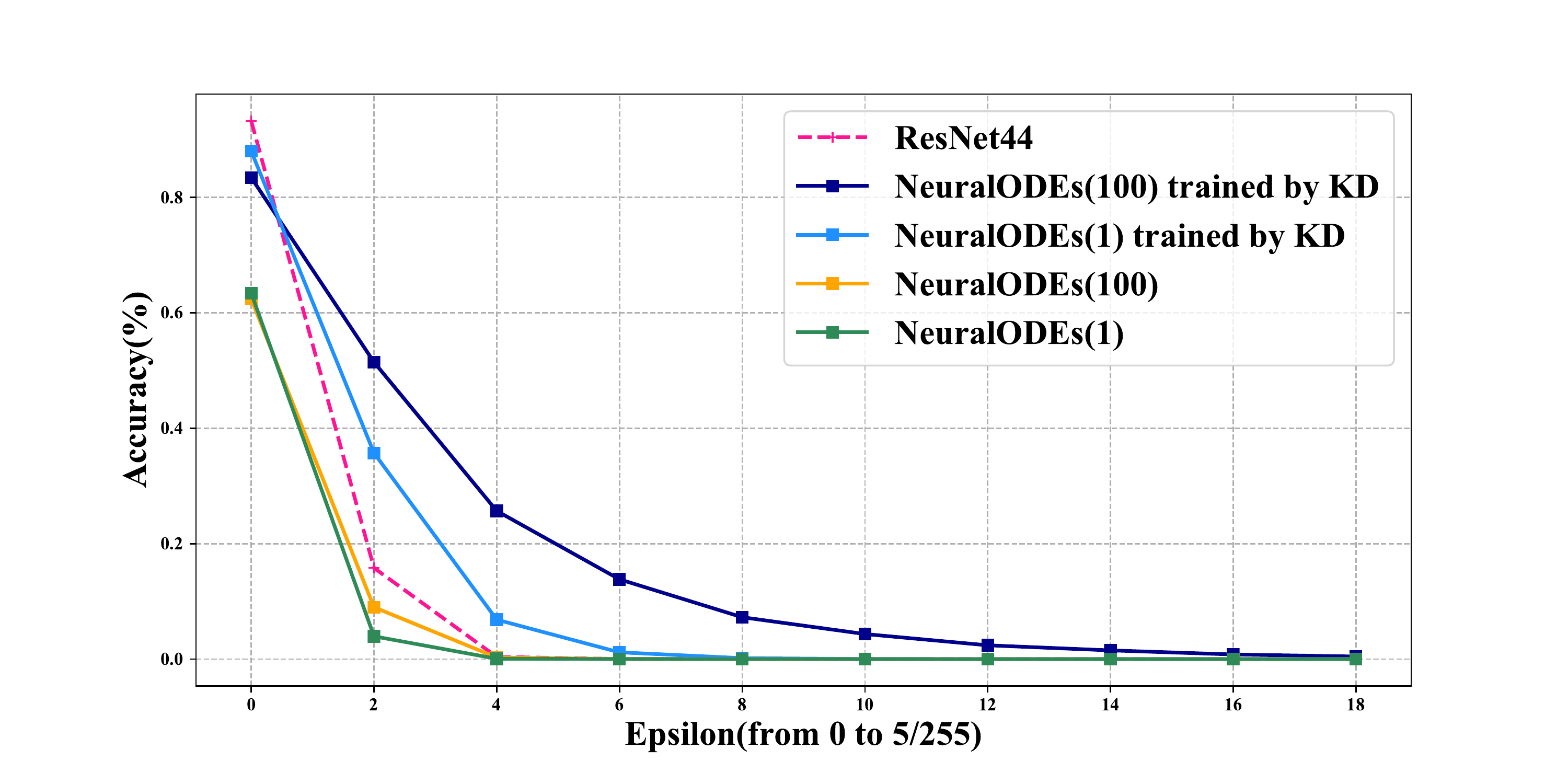}
 \label{fig:subfigure3}}
\quad
\subfigure[The evaluation on SVHN under MI-FGSM.]{%
\includegraphics[width=7.9cm]{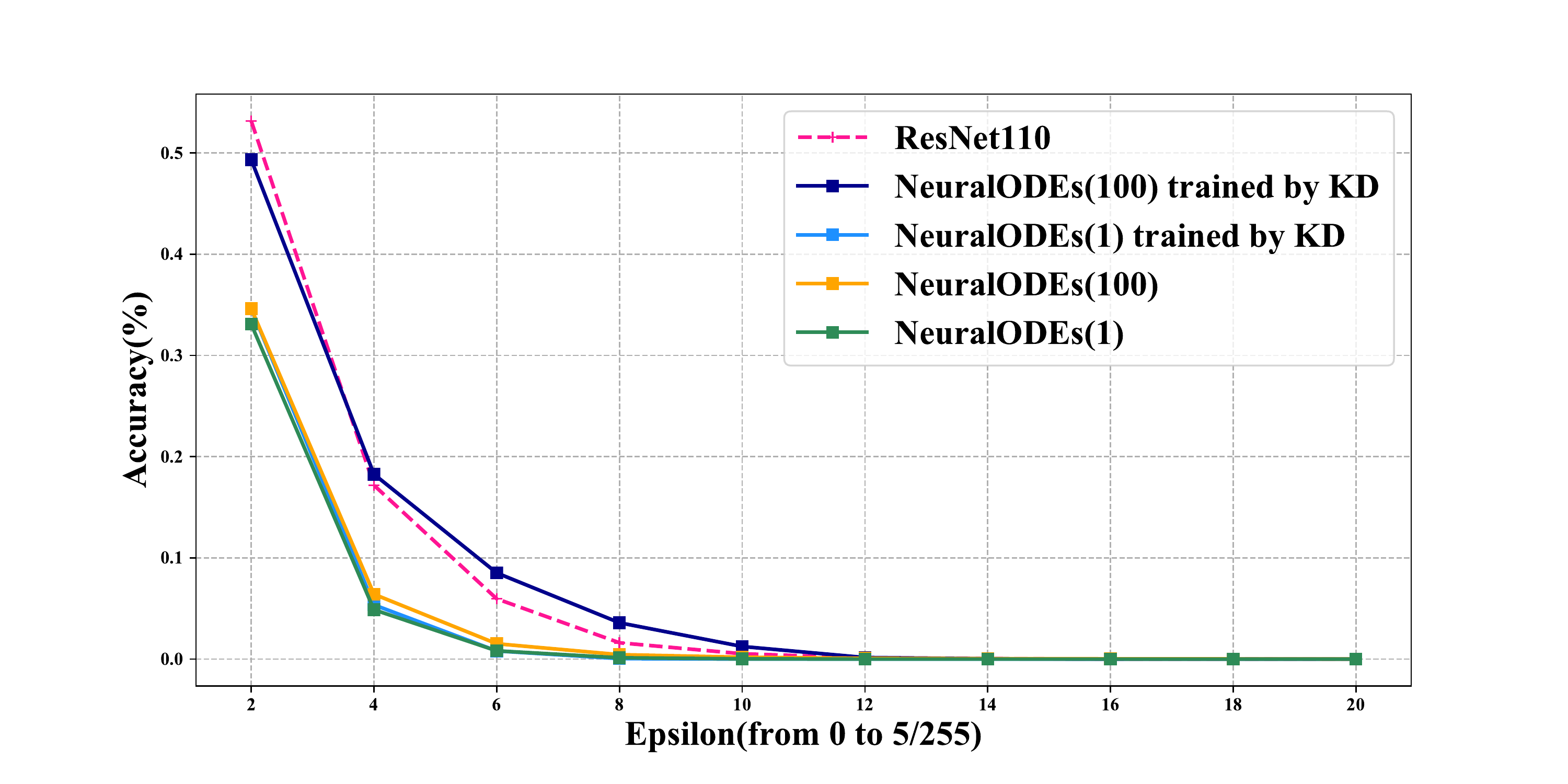}
  \label{fig:subfigure4}}
\end{center}
\caption{The classification accuracy of Neural ODEs trained by knowledge distillation on CIFAR10 and SVHN under three white-box attacks.}
\label{fig:figure}
\end{figure*}

\begin{table*}[t]
\caption{The classification accuracy of Neural ODEs(500) without using knowledge distillation on CIFAR10 under PGD.}
\centering
\begin{tabular}{ccccccc}
\toprule
Datasets & Model & clean & $\epsilon=8/255$ & $\epsilon=12/255$ & $\epsilon=16/255$ & $\epsilon=20/255$ \\
\midrule
CIFAR10 & Neural ODEs(500) & 69.94 & 25.07 & 23.60 & 21.29 & 20.24  \\
\bottomrule
\end{tabular}
\label{exp:node500_cifar}
\end{table*}

\subsection{The Training of Neural ODEs without Using Knowledge Distillation}

As shown in Table~\ref{tab:node_resnet_original}, the classification accuracy of Neural ODEs on CIFAR10 and SVHN is lower than the state-of-the-art neural networks without using knowledge distillation. Increasing the time horizon means increasing the time for solving ODEs, which will slow down the training speed of the Neural ODEs. However, we surprisingly find that increasing the time horizon can remarkably increase the robustness of Neural ODEs against adversarial examples, which will be described in detail in Section~\ref{sec:time_horizon}. 

\subsection{The Training of Neural ODEs Using Knowledge Distillation}

The classification accuracy of the five teacher ResNet models on CIFAR10 and SVHN test sets is shown in Table~\ref{tab:node_resnet_original}. Experimental results show that the new training manner can improve the classification accuracy of Neural ODEs by 24.58\% on CIFAR10 and 5.17\% on SVHN. With knowledge distillation, Neural ODEs can achieve competitive results on CIFAR10 and SVHN test sets compared with ResNet. We fine-tune the temperature $T$ in Eqn.\ref{equ:kd}. The results are summarized in Table~\ref{exp:kd_finetune_t}. 

Although none of the student Neural ODEs models outperform the teacher ResNet models in terms of classification accuracy on CIFAR10, some student models outperform the teacher models on SVHN, which is a good indication of the learning capability of Neural ODEs on image recognition tasks. As long as sufficient supervised information is provided, the performance of Neural ODEs can be comparable to that of the discrete version of deep neural networks.

\subsubsection{Effect of Temperature on Knowledge Distillation}

Using a high value for temperature $T$ produces a softer probability distribution over classes. For CIFAR10, the distillation results at a small temperature are inferior to those at a large temperature. In contrast, the improvement in knowledge distillation is limited when the temperature is increased to a certain value. For SVHN, good distillation results are achieved when the temperature is small, and increasing the temperature decreases the distillation effect.

\subsubsection{Effect of the Teacher Model on Knowledge Distillation}

Mirzadeh et al.~\cite{mirzadeh2020improved} claimed that the gap between student and teacher models is a key to the efficacy of knowledge distillation, and the student model performance may decrease when the gap is larger. So it is an interesting question whether an optimal teacher ResNet for Neural ODEs can be found. As mentioned above, it is unclear how to define the depth of the Neural ODEs. Thus, it is difficult to define the gap between the student Neural ODEs model and teacher ResNet models. In our experiments, we attempt to use five different kinds of ResNets to find the relationship between the number of layers of ResNet and the distillation result. Still, unfortunately, the experimental results do not present a particular pattern. 

\subsection{The Robustness of Neural ODEs against Adversarial Examples} \label{sec:robustness} 

We test the performance of the original Neural ODEs and the Neural ODEs trained by knowledge distillation on two white-box adversarial attacks: PGD~\cite{madry2017towards} and MI-FGSM~\cite{dong2017discovering}. 

\subsubsection{The Configurations of Adversarial Attacks} 

\textbf{PGD} \ We set the size of perturbation $\epsilon$ of PGD in the infinite norm sense. The pixel range is set to [0, 1]. The step size is set to 1/255, and a uniform random perturbation is added before performing PGD. 

\textbf{MI-FGSM}\ We set the pixel range to [0, 1], the step size to 1/255, and the momentum factor $\mu$ to 1.

For both PGD and MI-FGSM, the size of perturbation $\epsilon$ is from 0 to 20/255. The number of steps $s$ is calculated as 
$s = \lfloor\text{min}(\epsilon \cdot 255 + 4, \epsilon \cdot 255 \cdot 1.25)\rfloor$.

\subsubsection{Knowledge Distillation Improves the Robustness of Neural ODEs}

As can be seen in Figure~\ref{fig:subfigure1} and~\ref{fig:subfigure3}, compared with the original Neural ODEs, knowledge distillation not only improves the accuracy of Neural ODEs on the CIFAR10 test set but also improves the robustness of Neural ODEs against adversarial examples generated by PGD and MI-FGSM. As the size of perturbation $\epsilon$ of PGD and MI-FGSM increases, the original Neural ODEs rapidly lose their resistance to the adversarial examples while Neural ODEs trained by knowledge distillation lose their resistance lower than the original Neural ODEs. Figure~\ref{fig:subfigure2} and~\ref{fig:subfigure4} show that knowledge distillation also improve the robustness of Neural ODEs on SVHN.

\subsubsection{Increasing the Time Horizon Improves the Robustness of Neural ODEs} \label{sec:time_horizon}

Without knowledge distillation, Neural ODEs(1) are not even as robust to adversarial examples as ResNet. Those experimental results show that the robustness of Neural ODEs is not from obfuscated gradients claimed by~\cite{huang2020adversarial}. Although the original Neural ODEs(100) are not as robust to adversarial examples as ResNet, Neural ODEs(100) trained by knowledge distillation are significantly robust to adversarial samples on CIFAR10. The resistance of Neural ODEs(100) trained by knowledge distillation persists until $\epsilon$ is 20/255, while the resistance of Neural ODEs(1) trained by knowledge distillation persists only until $\epsilon$ is 8/255. The robustness of Neural ODEs(100) trained by knowledge distillation on SVHN does not outperform that of ResNet, but increasing the time horizon still improved the robustness of Neural ODEs. Nevertheless, the distillation effect slightly decreased when increasing the time horizon. 

We expect Neural ODEs with a larger time horizon will be more robust to adversarial examples. However, the training time of Neural ODEs with a large time horizon will be longer than that of Neural ODEs with a small time horizon. Because testing Neural ODEs is extremely time-consuming, we train Neural ODEs with the time horizon [0, 500] and only test it on PGD to show that increasing the time horizon can significantly improve the robustness of Neural ODEs against adversarial examples. The experimental results are summarized in Table~\ref{exp:node500_cifar}. The experimental configurations of Neural ODEs(500) can be seen in the supplementary material. As one can see, Neural ODEs(500) are very robust to adversarial examples. Even when $\epsilon$ of PGD is 20/255, the classification accuracy of Neural ODEs(500) remains at 20\%. These experimental results verified that the robustness of Neural ODEs benefits from the time horizon rather than obfuscated gradients.

\section{Conclusion}

In this paper, we analyzed that the possible reason for the poor performance of Neural ODEs on image recognition tasks is insufficient supervised information. Therefore, we proposed a new training manner based on knowledge distillation to alleviate this problem. Specially, we treated ResNet as the teacher model to provide richer supervised information. The experimental results showed that knowledge distillation improves the accuracy of Neural ODEs on CIFAR10 by 24\% and on SVHN by 5\%. The accuracy of the student Neural ODEs model even surpassed that of the teacher ResNet model on SVHN. In addition, we found that knowledge distillation can improve the robustness of Neural ODEs against adversarial examples. More importantly, we found that the robustness of Neural ODEs benefits from the time horizon rather than obfuscated gradients. Increasing the time horizon can remarkably improve the robustness of Neural ODEs. However, there is a trade-off that increasing the time horizon slows down the training of Neural ODEs and slightly reduces the effect of knowledge distillation.

\bibliographystyle{IEEEbib}
\bibliography{icme2022template}

\end{document}